\documentclass[sigconf]{acmart}

\acmConference[ESEC/FSE 2023]{The 31st ACM Joint European Software Engineering Conference and Symposium on the Foundations of Software Engineering}{11 - 17 November, 2023}{San Francisco, USA}
\usepackage[utf8]{inputenc}
\usepackage{graphicx}
\usepackage{multirow}
\usepackage{multicol}
\usepackage{amsmath}
\usepackage{enumitem} 

\DeclareMathOperator*{\argmax}{argmax} 

\title{Automatic Generation of Attention Rules For Containment of Machine Learning Model Errors}
\author{Samuel Ackerman, Axel Bendavid,  Eitan Farchi, Orna Raz}
\date{November 2022}

\begin{abstract}
Machine learning (ML) solutions are prevalent in many applications.
However, many challenges exist in making these solutions business-grade. For instance, maintaining the error rate of the underlying ML models at an acceptably low level.
Typically, the true relationship between feature inputs and the target feature to be predicted is uncertain, and hence statistical in nature.
The approach we propose is to separate the observations that are the most likely to be predicted incorrectly into `attention sets'. 
These can directly aid model diagnosis and improvement, and be used to decide on alternative courses of action for these problematic observations.  
We present several algorithms (`strategies') for determining optimal rules to separate these observations.  In particular, we prefer strategies that use feature-based slicing because they are human-interpretable, model-agnostic, and require minimal supplementary inputs or knowledge. In addition, we show that these strategies outperform several common baselines, such as selecting observations with prediction confidence below a threshold.
To evaluate strategies, we introduce metrics to measure various desired qualities, such as their performance, stability, and generalizability to unseen data; the strategies are evaluated on several publicly-available datasets. 
We use TOPSIS, a Multiple Criteria Decision Making method, to aggregate these metrics into a single quality score for each strategy, to allow comparison.  

\end{abstract}

\begin{document}

\newcommand{\build}[1]{\ensuremath{{#1}_{\textrm{build}}}}
\newcommand{\eval}[1]{\ensuremath{{#1}_{\textrm{eval}}}}
\newcommand{\train}[1]{\ensuremath{{#1}_{\textrm{train}}}}
\newcommand{\test}[1]{\ensuremath{{#1}_{\textrm{test}}}}

\newcommand{\nth}[1]{#1^{\text{th}}}
\newcommand{\splitatcommas}[1]{\begingroup\lccode`~=`, \lowercase{\endgroup
    \edef~{\mathchar\the\mathcode`, \penalty0 \noexpand\hspace{0pt plus 1em}}%
  }\mathcode`,="8000 #1%
  }
\newcommand{\splitatampersand}[1]{\begingroup\lccode`~=`& \lowercase{\endgroup
    \edef~{\mathchar\the\mathcode`& \penalty0 \noexpand\hspace{0pt plus 1em}}%
  }\mathcode`&="8000 #1%
  }

\maketitle

\section{Introduction
\label{sec:intro}
}

One common challenge of deploying and maintaining machine learning (ML) solutions is providing useful diagnoses of the model performance on datasets.  Here, we consider only ML classifiers rather than numeric predictors; we also consider data in the form of structured tabular datasets, or items from which structured meta-features may be extracted.  That is, let $\mathbf{X}$ be a tabular matrix of $p$ feature columns; $Y$ be a corresponding categorical-valued vector of ground truth label values; and $\hat{Y}$ be the vector of predicted values of $Y$, output by a trained ML classifier model $\mathcal{M}$. 

A classifier $\mathcal{M}$ models the unknown true relationship $f\colon\:\mathbf{x}\rightarrow y$ between vector-valued input $\mathbf{x}$ and the target $y$.  Hence, for a particular input $\mathbf{x}_i$, while it is unknown if $\mathcal{M}$'s prediction will be an error (i.e., $y_i\ne\hat{y}_i$), some values are more likely to be mis-classified, and we can model the likelihood of this. In this work, we address model diagnosis by providing the user with an `\textbf{attention set}', a set of dataset observations on which the classifier model's predictions are particularly likely to be errors. The attention set serves to localize the classification errors, by containing as many (likely) errors as possible and as few (likely) correct predictions.
A `\textbf{strategy}' is an algorithm that selects the attention set.  The selection may be done by the intermediate step of determining an `\textbf{attention rule}', typically by some optimization procedure.  An attention rule is a rule or criterion that can create an attention set by collecting all observations in the dataset that satisfy the rule.

A `\textbf{build}' dataset, denoted \build{D},
is the dataset on which a given strategy determines an attention rule.  $\build{D}$ must consist of $(\mathbf{X},Y,\hat{Y})$, or possibly only $(\mathbf{X},Z)$, where $Z=I(Y=\hat{Y})$; that is, we must know which observations are misclassified by $\mathcal{M}$.
Typically, it is most useful if a strategy (see Section~\ref{ssec:strategies}) defines attention rules that are human-interpretable and deterministic.  As part of the diagnosis, the user may receive both the attention set and the rule.
The attention rule should deterministically select the attention set in a way that can be mapped to an unseen `\textbf{evaluation}' dataset \eval{D} on which we do not know if observations were classified correctly (i.e., $Y$ is unknown, only its $\mathbf{X}$ is available).


In particular, if we assume that any given build or evaluation dataset represent identically-distributed samples from a hypothetical `population' dataset, an attention rule should generalize well \textit{statistically} to an unseen evaluation dataset in the sense that the attention set should have similar measured properties (e.g., size as a fraction of the dataset size) on \eval{D} as on \build{D}.  In addition, a good rule should be related to the likelihood of an observation being misclassified, since that is the localization aim.    Consider, for instance, a rule to form the attention set from a random sample of 5\% of observations in \build{D}.  While it may have similar statistical attributes on \eval{D}, based on the observations being randomly sampled, the rule criterion will not be \textit{effective} in the localization because random sampling does not target errors with higher likelihood than non-errors.  Thus, the strategy is an algorithm to select an optimal attention set, typically by way of specifying an intermediate optimal attention rule.  Since this is a predictive task, an attention set on $\eval{D}$ (and often on $\build{D}$) will typically not have 100\% error concentration.

Given the attention set selected by a given strategy, a user may have a given procedure (which we term a `\textbf{policy}') for performing diagnosis on the model.  A policy may also entail subjecting the attention set to differential treatment to obtain a value for $\hat{y}_i$ other than that predicted.  Several potential policies are
\begin{itemize}[leftmargin=*]
    \item \textbf{Diagnosis}: Manually examine attention set provided by the strategy on \build{D} and infer ways to improve the ML model (e.g., which feature values cause observations to be difficult to classify).  If the attention rule is defined by feature-based slicing, one can determine which features used in the definition were most responsible for the errors, such as by automatic methods such as SHAP values (\cite{LL2017}).
    \item \textbf{Treatment}: Each attention set observation on \eval{D} is routed to a human to judge whether the model prediction is correct or not (i.e., human in the loop).
    \item \textbf{Treatment}: 
    Implement some default non-model-based prediction rule for attention set observations on \eval{D}, without human routing.    For instance, set the label as a constant hard-coded value or hard-coded if-then or similar rule.    
   
\end{itemize}

\begin{figure}[h]
    \includegraphics[scale=1.0]{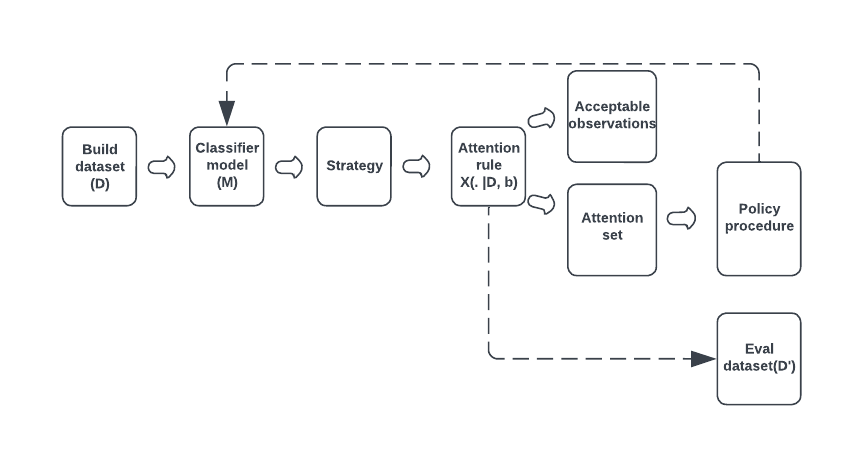}
    \caption{Schematic of a strategy within a policy
    \label{fig:strategy_policy_budget}}
\end{figure}

To be useful or manageable\footnote{Because an ML model should handle a large enough portion on the inputs to save human effort and because of practical time and effort constraints on the part of a human examiner when an input is highlighted as potentially mis-classified by the ML model.}, a strategy should be able to find a relatively small attention set (e.g., no more than 10\% of the dataset) with a concentration of errors that is high relative to that in the dataset on average.  We call this size restriction a `\textbf{budget}'; by construction, the budget is an upper bound on (empirical) probability of an observation from \build{D} being in the attention set, which hopefully should be similar on \eval{D}. Figure~\ref{fig:strategy_policy_budget} illustrates the relationship between a strategy, a policy, and the budget.

In this work we formulate several strategies and propose quality metrics for evaluating them.  
In particular, several strategies are based on the output of FreaAI (\cite{ARZ2020}) technology, which outputs statistically-significant and interpretable error-concentrated slice rules based on the values of prediction features (see Section~\ref{sec:FreaAI_background}), which are useful in other applications (e.g., \cite{ADFRZ2021}).  In principle, however, the slice rules could also come from other sources, such as human manual specification.  The slice-based strategies create attention sets formed by unions of these slices to optimize mis-classification concentration and coverage given a budget. As a baseline for our evaluations, we formulate strategies to select the attention set by finding observations with low values of the classifier confidence, or with ground truth label values that are identified as having low accuracy. 

This work is restricted to formulating and evaluating various strategies.  We do not conduct experiments to determine the effectiveness of various policies, that is, what to ultimately do with the attention sets found by the strategies.  However, we believe that strategies based on the FreaAI feature-value slices are well-suited to any policy procedure.
As discussed in Section~\ref{subsub:strategies_overview}, this is because such slices are more human-interpretable than criteria used by other strategies, and may also generalize better to unseen datasets.  Furthermore, rules in slice form are easier to understand and can be used directly in model diagnosis.  We found that the most successful strategy overall---as determined by the TOPSIS aggregation of the strategy quality metrics---uses FreaAI slices in an adaptive set cover algorithm to create the attention rule by selecting slices that optimally cover observations not contained in other slices.

An outline of this work is as follows: Section~\ref{sec:FreaAI_background} provides relevant background on the FreaAI slice-finding technology used in some of the strategies.  Section~\ref{sec:methodology} introduces general notation for strategies. Section~\ref{sec:strategy_metrics} uses the notation of Section~\ref{sec:methodology} to introduce several quality metrics to evaluate the success of a given strategy on a dataset.  Section~\ref{sec:experiment} describes the setup of our experiments to compare the strategies across different datasets. Section~\ref{sec:results} describes the results of the experiments.  Section~\ref{sec:related} describes related work and compares them to our method.  Section~\ref{sec:conclusion} concludes.

\section{{Frea-AI} Background
\label{sec:FreaAI_background}}

FreaAI (\cite{ARZ2020}) is a technology that implements a set of heuristics to efficiently suggest data slices (specific combinations of values of prediction features) that localize observations that have been incorrectly classified by a machine learning predictor.  Individually, these slices are, by construction, human-explainable and statistically significant. Each slice also has a real-valued rank $\in[0,1]$, where a higher value indicates better capturing of errors\footnote{The rank balances trade-offs between various slice properties.  For instance, a slice with a larger observation support (better localization) on average will have a lower error concentration (worse localization).  The rank is calculated by a smooth polynomial fit on these two properties across all slices found on the dataset.  An alternative ranking is presented in \cite{FNN2021}.}. Furthermore, each slice typically contains only a small proportion of the misclassified observations in the dataset, but together they form a sort of diagnostic report on the classifier's performance.  Slices often overlap in the observations they cover. 

We now describe how FreaAI slices are found (see \cite{ARZ2020}). 
Assume a model $\mathcal{M}$ is trained on a dataset $D$ with feature matrix $\mathbf{X}=\begin{bmatrix}F_1,\dots, F_p\end{bmatrix}$ and a target feature $Y$; the ML model returns prediction vector $\hat{Y}$. A data
slice $S$ on the dataset $D$ is defined in terms of the feature
space of $D$ (ignoring the target $Y$). $S$ is a rule indicating
value ranges for numeric features, sets of discrete values for categorical features, and combinations of the above. One hypothetical slice is $S_1 = (\textrm{RACE} \in \{\textrm{Black},\textrm{White}\})\: \& \:(10 \leq \textrm{EXPERIENCE} \leq 13)$, an intersection
of subsets of the two features ‘RACE’ and ‘EXPERIENCE’.  Slices are formed in this way because the conjunction of a low number of feature subsets is inherently `interpretable'; for instance, one can intuitively understand that slice $S_1$ contains people of particular race and who have a particular range of work experience.

Although FreaAI works only with model classification results on structured tabular data, it can be adapted to non-tabular data such as free text or images. For such data types, structured meta-features---such as the image size or contrast value, or text string length or number of dictionary words---can be automatically extracted into a tabular dataset, and FreaAI applied as usual.

Slices help users understand for which feature-value combinations the model has relatively high error (is `weak'), and are found by the following procedure:
\begin{enumerate}[leftmargin=*]
    \item Form the binary target $Z$, where $z_i=I(y_i=\hat{y}_i)$.  $z_i$ indicates whether observation $i$ is classified correctly or not.
    \item An exhaustive iteration is performed on all unique unordered combinations (typically only of sizes 1, 2, or 3) of features $\{F_1,\dots,F_p\}$.
    \item For a given combination, say $\{F_1,F_3\}$, a decision tree model is trained on $D$ to predict $Z$.
    \item Say $\zeta$ is the set of observation indices contained in a tree node.  The node support is $|\zeta|$ and, accuracy is $\frac{\sum_{i\in\zeta}z_i}{|\zeta|}$; the error rate, also known as the misclassification rate (MCR), is $1-\textrm{accuracy}$.
    \item The slices are formed by taking the terminal tree leaves that satisfy constraints on minimum support (so slices are large enough so they represent a significant aspect and not chance noise or artefacts) and maximal accuracy (at a threshold significantly lower than the average to ensure they contain significant error concentration).
    \item Statistical significance is determined by a lower-tailed hypergeometric test (see \cite{ARZ2020}), and slices must have a p-value lower than 0.01.
\end{enumerate}

The original goal of the FreaAI analysis in \cite{ARZ2020} was that the list of these slices would be used to perform corrective action on the ML model, such as by automatic means or by manual examination by a human (i.e., a `policy').  
However, even though each slice individually contains a statistically-significant error concentration, practically using this list is difficult for several reasons.  First, there may be many (e.g., several thousand) such slices found.  Second, the union of the observations contained in each slice often covers a high proportions of the observations in the dataset $D$; hence a simple union is impractical for error localization, since it will likely exceed a reasonable budget for the level of effort required to analyze it.  Furthermore, even though each slice is human-interpretable (because it is defined on feature values), the full union will likely not be.  These factors motivate this work, in which we find several strategies to create attention sets by optimal selection of some subset of the slices, for instance by considering their overlaps in observation coverage, or by minimizing the number of features used in the slices forming the attention set so that the result is still reasonably simple for a human to understand.

In addition to human-interpretability, creating attention sets from the union of selected FreaAI slice rules has desirable properties that we define and measure next, in Sections \ref{sec:methodology}
and \ref{sec:strategy_metrics}.



\section{Methodology
\label{sec:methodology}}




Section~\ref{sec:attention_set} introduces basic notations and statistics with regard to attention sets. Section~\ref{ssec:strategies} formulates the strategies discussed (FreaAI-based ones vs traditional ones).
Section ~\ref{ssec:guidelines} shares common strategy guidelines.

\subsection{Attention set
\label{sec:attention_set}}
An attention set is
the collection of all observations in a dataset satisfying a given attention rule.  Let $X_{\textrm{strategy}}(\cdot \mid \build{D}, b)$ represent the fitting of the strategy or attention rule to a build dataset \build{D}, given a maximal budget $0<b\leq 1$ constraint on its size.  For fixed choices of $b$, \build{D}, and a strategy algorithm, we denote it for clarity by $X(\cdot)$.  $X(\cdot)$ is a deterministic function that implicitly contains the attention rule; it receives an input dataset $D$ (possibly \build{D} itself) and returns the attention set resulting from applying the attention rule fit on \build{D}. The output $X(D)\subseteq D$ returns an attention set that is a subset of $D$.  For instance, if the optimal attention rule satisfying the budget $b$ is $X(\cdot)=$``all observations in `$\cdot$' where $\textrm{AGE}\geq 60$" or ``all observations in `$\cdot$' where the classifier confidence is $\leq0.2$", then $X(D)$ returns the attention set on $D$, that is, all observations in $D$ satisfying this criterion.  Note, $X(\build{D})$ must be equivalent to the attention set on \build{D} used to define $X(\cdot)$ in the first place, which by construction must contain at least one observation.  Hence $X(\build{D})\ne \emptyset$, but on other inputs $D\ne\build{D}$, we may have $X(D)=\emptyset$.

Practically, we are only interested in $b\leq B$, where $B$ is some maximal budget coverage value representing practical constraints. If $B=0.2$, for example, then we only find attention sets representing up to at most 20\% of the observations on \build{D}.

Let $D$ denote any dataset, either \build{D} or \eval{D}.
The metrics defined below can be applied equally to any $D$, though other than $N(\cdot)$, they require $D$ to have the true labels $Y$, which we retain in our experimental setup but may not have in practice.

\begin{itemize}[leftmargin=*]
    \item Let the function $N(\cdot)$ denote the size or support of its input $\cdot$.  In particular,
        \begin{itemize}
            \item Let $N(D)\geq 1$ be the number of observations in the dataset $D$, and 
            \item Let $N(X(D))$ be the size of the attention set selected by the attention rule $X(\cdot)$ on the dataset $D$. 
            Because the attention rule always defines a proper subset on the dataset they are applied to, $N(X(D))\leq N(D)$.
            \item If $N(X(\build{D}))=0$, no build attention set was found satisfying the budget $b$, that is, no rule is defined.  Then, $N(X(\eval{D}))=0$ by definition since there is no rule.  All the statistics should be considered undefined rather than zero-valued.
            \item If $N(X(\build{D}))\geq 1$ (a build rule was found satisfying $b$), then it's possible that $N(X(\eval{D}))=0$.
        \end{itemize}
        \item Let $\bar{N}(X( D))=\frac{N(X(D))}{N(D)}$ be the fractional size of the attention set selected by the strategy $X(\cdot)$ on the dataset $D$.
        \begin{itemize}
            \item If $1\leq N(X(\build{D}))$, then $0<\bar{N}(X(\build{D}))\leq b$, because the build attention set must satisfy the budget $b$.
            \item However, since $X(\cdot)$ may exceed the budget on \eval{D}, 
            then $0 \leq \bar{N}(X(\eval{D}))\leq 1$. A stable strategy should have $\bar{N}(X(\eval{D}))\approx \bar{N}(X(\build{D}))$.
        \end{itemize}
        
        \item Let $M(D)$ and $M(X(D))$ be the number of misclassified observations in $D$ and in the attention set, respectively.  That is, $M(D)=\sum_{i=1}^{N(D)}I(y_i\ne \hat{y}_i)$ and $M(X(D))=\sum_{i\in X(D)}I(y_i\ne \hat{y}_i)$.
    \end{itemize}

We define the following statistics using the above notation.  Table~\ref{tab:notation_illustration} illustrates the statistics for two budgets.

\begin{figure}[h]
    \centering
    \includegraphics[width=\columnwidth]    {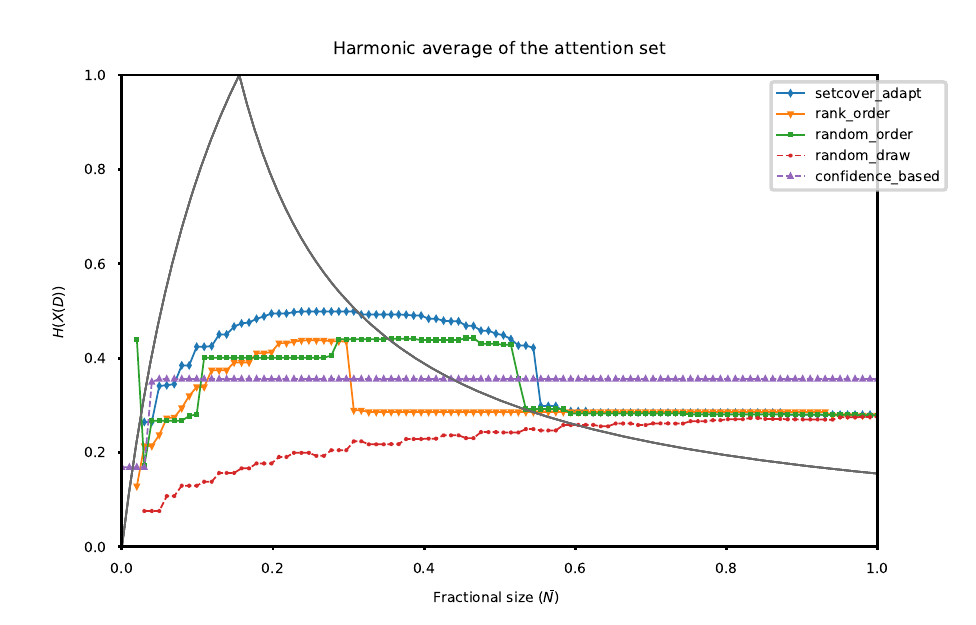}
    \caption{
    Harmonic mean $H$ (Section~\ref{sec:attention_set}) for strategies on the Adult dataset, averaged across build splits, plotted vs the attention set fractional size $\hat{N}$.  The black line shows the best possible value of $H$ on the full Adult dataset (MCR=0.15) at each budget $b$, which occurs if an attention set of fractional size $\bar{N}\leq MCR$ contains only errors, and if $\bar{N}>MCR$, it contains all errors ($MC=1$), which maximizes the error rate $\bar{M}$, and thus $H$.  $H$ is maximum at $\bar{N}=MCR$ (here, 0.15),  which lends support to the idea that the maximum observed $H$, if it is less than $B=0.20$, the maximum acceptable budget, can be used to indicate the best stopping budget for a strategy.  The observed $H$ for strategies other than the random draw and confidence-based follow a similar unimodal pattern, but tend to peak when $\bar{N}>B$ already.}
    \label{fig:harmonic_mean}
\end{figure}

\begin{enumerate}[leftmargin=*]
    \item \textbf{Error rate}:  $\bar{M}(X(D))=\frac{M(X( D))}{N(X(D))}$, the misclassification rate (MCR) among observations in the attention set.
    \item \textbf{Misclassification coverage}: $MC(X( D))=\frac{M(X(D))}{M(D)}$, the proportion of misclassifications in the dataset that are contained in the attention set. So for instance, regardless of the dataset error rate $\bar{M}(D)=\frac{M(D)}{N(D)}$, a random subset of fractional size $b$ should cover $b$ proportion of $D$'s misclassifed observations, though each observation has a different probability of being an error depending on the error rate.
\end{enumerate}
\noindent
These first two statistics reflect the fact that a good attention should both cover many errors (high $MC$) relative to its size, that as many of its observations as possible should be errors (high $\bar{M}$).  Hence, these statistics are central in defining the other statistics and quality metrics.
\newline
\begin{enumerate}[leftmargin=*]
    \setcounter{enumi}{2}
    \item  \textbf{Harmonic average between the error rate and the misclassification coverage}: $H(X(D))=\frac{2*MC(X( D))*\bar{M}(X(D))}{MC(X( D)) +\bar{M}(X(D))} \in [0,1]$.  Similarly to how the F-1 score of a classifier is the harmonic mean of its precision and recall scores, $H$ balances the competing qualities of completeness ($MC$ error coverage) and the homogeneity ($\bar{M}$ error rate).  It can allow us to choose between different attention sets, such as for the two budgets in Table~\ref{tab:notation_illustration}, where $H$ is higher for $b=0.05$, indicating this may be the preferred attention set. This attention set when $b=0.05$ is smaller but has higher error rate compared to the larger set resulting from utilizing the entire budget $b=0.10$.  See Figure~\ref{fig:harmonic_mean} for discussion. 
\end{enumerate}

\noindent
If the policy procedure is to either discard the attention set entirely or correct the model based on these observations, we have two further relevant statistics:

\begin{enumerate}[leftmargin=*]
    \setcounter{enumi}{3}
    \item \textbf{Error rate in remaining observations}: $\bar{M}(D - X(D))=$\\$\frac{M(D) - M(X(D))}{N(D) - N(X(D))}$; that is a measure of the error rate in ``sanitized" data, i.e., the data after removing the attention set observations.
    \item \textbf{Error rate in full dataset if attention set errors are fixed}: \\$\frac{M(D) - M(X(D))}{N(D)}$, that is, the error rate if theoretically a human could fix the ML model so that all errors in the attention set would not exist, and return them to the original dataset.

    
\end{enumerate}


\begin{table}
\begin{tabular}{|p{4cm} | p{1.8cm} | p{1.8cm}|}
\cline{2-3}
\multicolumn{1}{c}{} & \multicolumn{2}{|c|}{Budget $b_i$}\\
\hline
Metric & 0.05 & 0.10\\\hline
Observations: $N(X( D))$ & 37 & 99\\
Fractional size: $\bar{N}(X(D))$ & 37/1000=0.037 & 99/1000=0.099 \\
Error count: $M(X(D))$ & 13 & 21\\
Error coverage: $MC(X(D))$ & 13/50=0.26 & 21/50=0.42\\
Error rate: $\bar{M}(X(D))$ & 13/37$\approx$0.351 & 21/99$\approx$0.212\\
Harmonic mean: $H(X(D))$ & 0.299 & 0.282\\
\hline
\end{tabular}
\caption{\label{tab:notation_illustration}Toy example of attention sets on a build dataset $D$ of size $N(D)=1000$ observations, with $M(D)=50$ classification errors, and thus an error rate of $\bar{M}(D)=\frac{M(D)}{N(D)}=0.05$.  Note, attention sets at all budgets $b_i$ should have an internal error rate of $\bar{M}(X(D))> 0.05$, since the goal is to have a higher-than-average error concentration.}

\end{table}

\subsection{Strategies
\label{ssec:strategies}}
The goal of this work is to find  attention rules defining attention sets that contain ML model errors in order to help diagnose the model. We consider the following six 
strategies for finding an attention set, given a budget $b$. Section \ref{subsub:FreaAIStrategy} defines three strategies based on FreaAI data slicing technology. Section \ref{subsub:TraditionalStrategy} defines three commonly-used strategies as a baseline to compare the best FreaAI strategy against. 

The strategies based on FreaAI slices do not assume any knowledge about the underlying ML classifier model, while the confidence based one, which is the commonly used 'traditional' technique, does. 
Further, the model confidence may not be available for a particular ML model or it may not correlate with model error.  

\subsubsection{FreaAI slices-based strategies} \label{subsub:FreaAIStrategy}
The first two strategies create an optimized union of FreaAI slices on \build{D} according to a greedy ordering of the slices. The third serves as a baseline for the first two.
\\\par
\begin{enumerate}[leftmargin=*]
    \item \textbf{Set-cover:} considers the set of the slices as a set-cover problem. It iteratively\footnote{At each iteration, a new candidate is added to the list, based on optimizing the ratio between its reward and cost.  We set the reward and cost to be the number of previously uncovered errors and non-errors, respectively, that the candidate slice would add to the existing union.} adds slices (the sets) to an ordered list, the union of which should optimally contain as many errors (the target items to be covered) as possible, while including as few as possible non-errors.  The iteration continues as long as the size of the union of observations (the attention set) does not exceed the budget $b$. Our experimental results in Section \ref{sec:results} indicate that this is the preferred algorithm.  We adapt the Python module \texttt{SetCoverPy} (\cite{Z2016}) by customizing its cost and reward functions. 
    \item \textbf{Rank order:} first order slices by their FreaAI rank, a heuristic aiming to reflect how good a slice is compared to the others (see Section~\ref{sec:FreaAI_background}), then take the union of slices, by their order, as long as the union size as a fraction of the dataset does not exceed the given budget $b$.
    \item \textbf{Random order:} similar to rank order, except that the slices are selected in a randomized order. It serves as a baseline for the first two FreaAI strategies, and is expected to result in worse attention sets (as defined in Section \ref{sec:attention_set}).  However, it should still contain above-average error concentration because each individual FreaAI slice does.
\end{enumerate}

In our experiments, for each slice-based strategies, we first determine
the slice ordering satisfying a maximum budget $B=0.20$.  Then, using the same slice ordering as for $B$, the attention sets for a sequence of $K$ budgets $0<b_1<b_2<\dots<b_K=B$ are determined for each $b_i$, as follows: we take the maximum number of slices from the beginning of the slice ordering, for which the union of their observations forms an attention set satisfying the budget $b_i$.

\subsubsection{Commonly-used baseline strategies}\label{subsub:TraditionalStrategy}
The first two strategies are often used for fault localization in ML models. The third strategy is a naive baseline strategy.

\begin{enumerate}[leftmargin=*]
    \setcounter{enumi}{3}
    \item \textbf{Worst-label filtering:} For each observed ground truth label $\ell$, calculate its error rate $\hat{M}(\ell)=\frac{\sum_iI(y_i=\ell\:\&\:\hat{y}_i\ne y_i)}{\sum_jI(y_j=\ell)}$, and rank them in descending order of $\hat{M}$, from worst to best, giving the order $\mathcal{L}=\{\ell_{(1)},\ell_{(2}),\dots\}$.  Then, determine the set $\overline{\ell}=\{\ell_{(i)}\}_{i=1}^t$ where $t=\argmax_k\{k\colon\: \left(\sum_{i=1}^k\sum_{j=1}^{N(\build{D})}I(y_j=\ell_{(i)}\right)\leq b\times N(\build{D})\}$, if it exists. That is, the attention rule is to take the most worst-accuracy labels as long as the number of observations with those label values does not exceed the budget $b$ on the \build{D}. One flaw of this strategy is that often, the rule $X(D)$ will not be defined; this happens if $t$ is undefined, that is, if observations with the lowest-accuracy label $\ell_{(1)}$ comprise more than fraction $b$ of dataset \build{D}.
    \item \textbf{Confidence-based:} Let $c_i$ be the confidence\footnote{For a given observation feature vector $\mathbf{x}_i$ , let the label ground truth and model $\mathcal{M}$-predicted values be $y$ and $\hat{y}_i$, respectively. Many classifiers determine the prediction by calculating some $p(y=\ell\mid \mathbf{x}_i)$, that is, the degree certainty in the correctness of each potential label value $\ell$ (e.g., softmax function) and then setting $\hat{y}_i=\argmax_{\ell} p(y=\ell\mid \mathbf{x}_i)$.  For scikit-learn \cite{scikit-learn} classifiers, this is accessed by the \texttt{predict\_proba} method.  The confidence assigned to the true label value, that is, $p(y_i\mid\mathbf{x}_i)$, could be used as a measure of confidence $c_i$, in that low values would be indicative of misclassifications, this cannot be used in evaluation datasets where the true $Y$ may not be observed.  Hence, we use $c_i=p(\hat{y}_i\mid\mathbf{x}_i)$, that is, the model's confidence in the label value it predicts, which is always known. However, other model-derived confidence metrics could be used as well.  See \cite{NCAB2022} for a discussion of various metrics of model confidence on observations.} of the classifier on \build{D} observation $\mathbf{x}_i,\:i\in 1,\dots,N(\build{D})$. Given a budget $b$, let $t=\argmax_c\{c:\:0< \frac{\sum_{i=1}^{N(\build{D})}I(c_i\leq c)}{N(\build{D})} \leq b\}$.  That is, the $\nth{(100b)}$ quantile of the confidence values $\{c_i\}$, accounting for ties, such that when mapped back to \build{D}, the attention set size is at most $b\times N(\build{D})$, that is, it is within the budget. If no such $t$ is found on \build{D} (e.g., the extreme case where all $\{c_i\}$ are equal, and hence the confidence is useless as a discriminator), the rule is not defined.
\end{enumerate}
Finally, we have a random baseline as a sanity check:
\begin{enumerate}[leftmargin=*]
    \setcounter{enumi}{5}
    \item \textbf{Random subsets:} takes a random set of $b\times N(D)$ observations without replacement. This strategy is highly naive as it does not target errors at all, and hence should perform the worst.
\end{enumerate}
    
\subsubsection{Strategies overview
\label{subsub:strategies_overview}}

The slice-based strategies have the advantage over the others that their rules are human-interpretable, since they construct (untransformed) `regions' of the dataset raw feature space.  Being human-interpretable means the rules can be used directly for manual diagnosis; for instance, if the slice $\{\textrm{AGE}\geq 60\}$ contains most of the \build{D} attention set observations, then $\mathcal{M}$ for some reason (perhaps due to correlated features) has trouble with this demographic.  Assuming there exists a real relationship $\mathbf{X}\rightarrow Z$, between the feature values and the likelihood of an error, a feature-based mapping based on \build{D} should tend to generalize better than others to unseen \eval{D}, because they directly target this relationship.  The confidence-based strategy is not human-interpretable because, say, knowing that "10\% of observations have $c\leq 0.3$" can help \textit{select} problematic observations but does not help understand what they have in common.  The worst-labels strategy is interpretable because it uses the label values $Y$; however, as Section~\ref{subsub:TraditionalStrategy} mentions, often it cannot return budget-satisfactory rules, as frequently happened in our experiments.  Although it is dataset-dependent, a strategy that frequently is unable to return an attention set satisfying the maximum budget $B$ will be less useful in practice.

All strategies require ground truth labels $Y$ on both \build{D} and evaluation \eval{D} to calculate the evaluation statistics from Section~\ref{sec:attention_set}, as well as to determine the \build{D} attention set rule in the first place, because we need to know $Z$, that is, which predictions are mistakes.  However, the slice-based strategies do not need to know $Y$ on \eval{D} (in many realistic scenarios these may be missing) to map the rules to \eval{D} because they are based solely on the feature values.  The confidence-based strategy also does not require $Y$ on \eval{D}, but it is not model-agnostic (whereas slice strategies are), in that it requires access to the trained model $\mathcal{M}$.  The worst-labels strategy needs to know $Y$ on \eval{D}.

\subsection{Strategy guidelines \label{ssec:guidelines}}

Attention sets across ascending budgets by construction grow in size, and cannot diminish in MC. Thus, given a strategy $s$ and two ascending budgets $b_i<b_j$ we have the following:

    \begin{itemize}
        \item Size (raw and fractional) increases: $N(X_{s}(D\mid D, b_i))\leq N(X_{s}(D\mid D, b_j))$ and $\bar{N}(X_{s}(D\mid D, b_i))\leq \bar{N}(X_{s}(D\mid D, b_j))$
        \item Misclassifications (raw total and fractional) increases: $M(X_{s}(D\mid D, b_i))\leq M(X_{s}(D\mid D, b_j))$ and $MC(X_{s}(D\mid D, b_i))\leq MC(X_{s}(D\mid D, b_j))$.
        \item Ideally, we would have $\bar{M}(X_{s}(D\mid D, b_i))\geq \bar{M}(X_{s}(D\mid D, b_j))$. That is, the attention set at low budgets would have a very high error rate, and then decrease. However, that is not necessarily so.
    \end{itemize}

\begin{figure}[h]
    \includegraphics[scale=1.0]
    {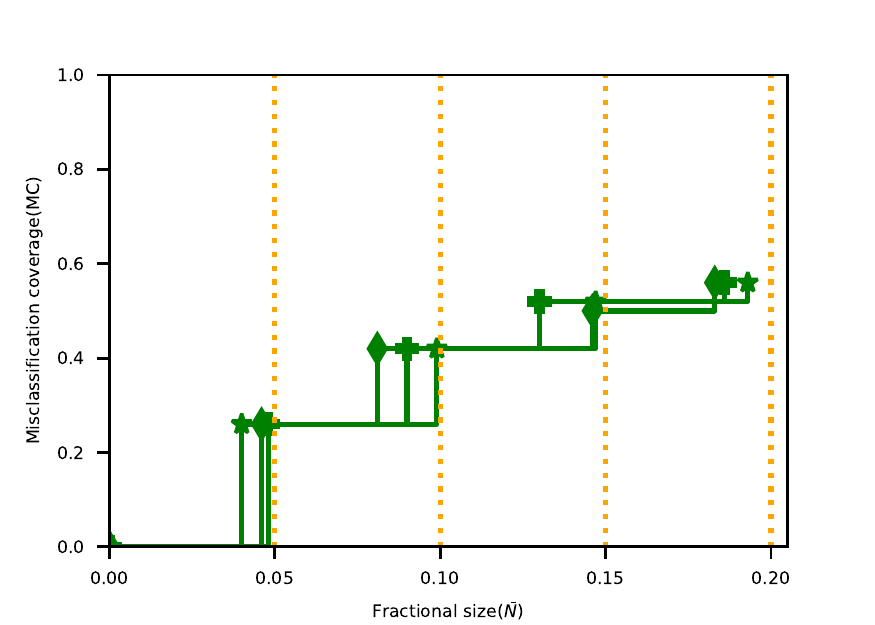}
    \caption{Toy example of step functions of attention sets selected at a set of 4 budgets increasing by a fixed $\delta=0.05$ given three different build dataset splits $\build{D}^1$, $\build{D}^2$ and $\build{D}^3 \subseteq \test{D}$ sharing the same feature space. The plot shows $MC(X_{s}(\build{D}^j\mid \build{D}^j, b_i))$ vs $\bar{N}(X_s(\build{D}^j\mid \build{D}^j, b_i)),\:j\in\{1,2,3\}$, across budget $b_i \in \{0.05, 0.10, 0.15, 0.20\}$.}    \label{fig:Attentionsets}
\end{figure}

Attention sets are typically constructed in a discrete manner, such as by adding slices to a union. Therefore, $N(X_{s}(D \mid \build{D}, b))$ typically increases in discrete jumps as $b$ increases continuously.  Our strategy evaluation metrics (Section~\ref{sec:strategy_metrics}) are largely based on diagrams of error coverage ($MC$) plotted on the vertical axis as a step function versus increasing fractional budgets $b$ on the horizontal axis.  Figure \ref{fig:Attentionsets} shows three toy examples of such step functions (indicated by marker symbols), where we see the $MC$ increase discretely at budges where the attention rule grows suddenly (as new slices added now satisfy the new budget).  We use plots of $MC$, rather than error rate $\bar{M}$, because they are more directly comparable across datasets with different MCRs $\bar{M}(D)$.  For instance, consider dataset $D_1$ and $D_2$ with MCRs $\bar{M}(D_1)=0.1$ and $\bar{M}(D_2)=0.01$. For a given budget, say $b=0.05$, for a given strategy $s$, we expect the attention set MCRs to be $\bar{M}(X_s(D_1))>\bar{M}(X_s(D_2))$ because of the difference in overall MCRs in the datasets.  But the misclassification coverage proportions $MC(X_s(D_1))$ and $MC(X_s(D_2))$ should be similar, and any differences should reflect the relative ease with which the strategy can discriminate errors (e.g., if a single small slice can contain many errors) in the datasets, but not the overall MCR.

\section {Strategy metrics \label{sec:strategy_metrics}}
To quantitatively assess the quality of a strategy we define the following desired properties that an attention set should have. To allow for comparison of different strategies we also define a single value (Section~\ref{subsec:topsis-score}) that combines these properties.

A strategy is assessed on a dataset $D$ by specifying a fine resolution of $K$ increasing budgets $0<b_1<b_2<\dots<b_K=B$, where, for instance, $b_{i+1}-b_i=\delta=0.01$ is a fixed increase of $\delta=0.01$ in the budget; we then find the strategy's attention set on $D$ at each $b_i$. 
A strategy is considered better if the attention sets found have the following properties:
\begin{enumerate}
    \item On a given \build{D}, they cover a high rate of misclassified inputs on average, across the increasing budgets. The \textbf{AUC} performance metric in Section  ~\ref{subsec:AUC} measures this property.
    \item They generalize well from \build{D} to \eval{D} in terms of similar misclassification coverage at all given budgets. The \textbf{generalizability} metric in Section~\ref{subsec:general}  measures this property.
    \item They are stable (have low variance) in terms of misclassification coverage at the same budget $b_i$ across multiple randomly sampled $\{\build{D}\}$.  The \textbf{stability} metric in Section~\ref{subsec:stability} measures this property.  See Section~\ref{subsec:procedure} for discussion of the sampling procedure.
\end{enumerate}

\subsection{Performance: AUC} \label{subsec:AUC}

\begin{figure}[h]
    \centering 
    \includegraphics[scale=1.0]
    {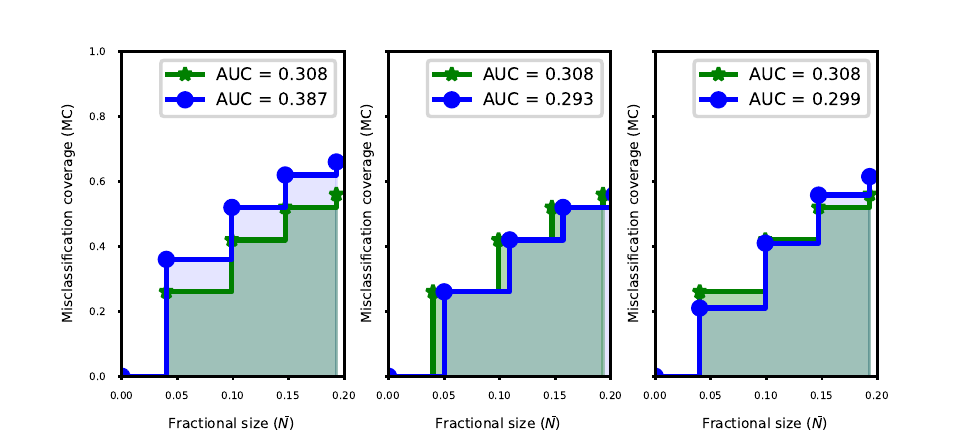}
    \caption{Comparison of AUC of two step functions.\\
    Left: Two strategies with the same span of the horizontal axis but where one succeeds in achieving a higher MC (vertically shifted) at all the budgets. \\
    Center: Horizontal shift so that the green has the same MC as the blue at each budget, but lower fractional attention set size, which is better.  The green will thus have a higher AUC.\\
    Right: Two different strategies can have the same AUC score.    
    }
    \label{fig:AUC_diagrams}
\end{figure}

Performance of a strategy is assessed by the area under the curve (AUC) of the error coverage step functions (see Figure~\ref{fig:AUC_diagrams}) on \build{D} and \eval{D}---denoted, respectively, $AUC_s(\build{D}\mid \build{D}))$ and  $AUC_s(\eval{D}\mid \build{D}))$---for a given strategy $s$.  We expect the same strategy $s$ to have roughly the same AUC on build datasets sampled from the same population, such as those in Figure~\ref{fig:Attentionsets}.

       
To formally define AUC on a dataset $D$, let us for clarity denote $m_i({D})=MC(X_{s}(D \mid \build{D}, b_i))$ and $n_i(D)=\bar{N}(X_{s}(D\mid \build{D}, b_i))$.  The ordered set of pairs $\{(n_i(D), m_i(D))\}_{i=1}^K$ define the misclassification coverage (MC) step function as shown in Figure~\ref{fig:Attentionsets}.
AUC of this step function on $D$ is defined as:
    \begin{equation}
        AUC_s(D\mid \build{D})=\frac{\sum_{i=1}^{K-1}m_i(D)\times(n_{i+1}(D)-n_{i}(D))}{n_K(D)}
    \end{equation}

Note that the AUC is normalized by the maximum attention set size $n_K$ reached, which includes both the span $[0, n_1]$ before the first budget $b_1$, and any $b_i>b_1$ where a rule could not be found.  As shown in the center plot of Figure~\ref{fig:AUC_diagrams}, this rewards strategies that are able to form attention sets earlier, at lower budgets.

       

\subsection{Generalizability} \label{subsec:general}
Generalizability captures the ability of the attention sets to generalize to unseen data, in terms of similar MC between the build and evaluation datasets, across the range of the relevant budgets. 

More formally, let us denote $e_i(\build{D}, \eval{D} \mid \build{D})=MC(X_{s}(\build{D}\mid \build{D}, b_i))-MC(X_{s}(\eval{D}\mid \build{D}, b_i))$, where $e_i\in(-1,1)$; that is, the MC  deviation between the attention sets selected on  \build{D} and \eval{D} by the same attention rule $X_s(\cdot \mid \build{D}, b_i$).  Then we can define a metric of generalizability to be:
\begin{equation}
   G_s(\build{D}, \eval{D}  \mid \build{D})=\frac{\sum_{i=1}^K \left(e_i(\build{D}, \eval{D} \mid \build{D})\right)^2}{\sum_{i=1}^K \left(m_i(\build{D})\right)^2}
\end{equation}

\begin{figure}[h]
            
               \centering \includegraphics[scale=1.0]
                {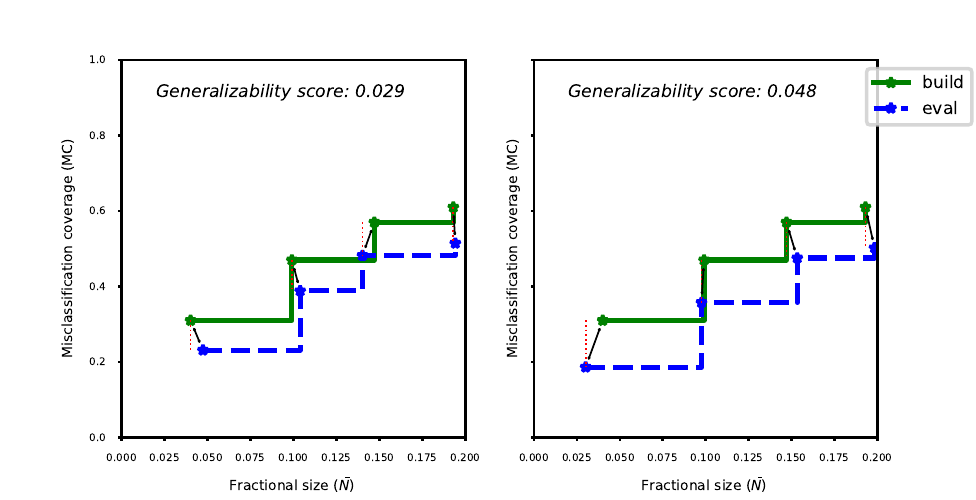}
                \caption{Generalizability scores of two strategies $s_1$ (left) and $s_2$ (right): $G_{s_1}(\build{D}^1, \eval{D}^1)$ and $G_{s_2}(\build{D}, \eval{D}^1)$. $s_1$ generalizes better than $s_2$. MC values for $s_1$ show smaller gaps between the build and evaluation sets.}
            \label{fig:gen}
    \end{figure}

$G_s$ is the variance-normalized mean squared error of the misclassification coverage, where higher values indicate less generalizability. Figure \ref{fig:gen} depicts the generalizability scores of two different strategies, where one (left) generalizes better than the other (right).

\subsection{Stability}
\label{subsec:stability}
Stability measures how consistent the building of an attention set is, given sampling variability of input data. To measure this, we generate
$Q>1$ independently and identically-distributed (IID) pairs   
$(\splitatcommas{\build{D}^1,\eval{D}^1}),\dots,(\build{D}^Q,\eval{D}^Q)$ from the source (`population') dataset $D$; $\build{D}^1,\dots,\build{D}^Q$ approximate the sampling variability of build datasets.  Given a strategy $s$, we fit $X_s(\cdot\mid \build{D}^j, b_i)$ for each combination of $j=1,\dots,Q$ and budget $b_i, \:i=1,\dots,K$.

\begin{figure}[h]
   \centering \includegraphics[scale=1.0]
    {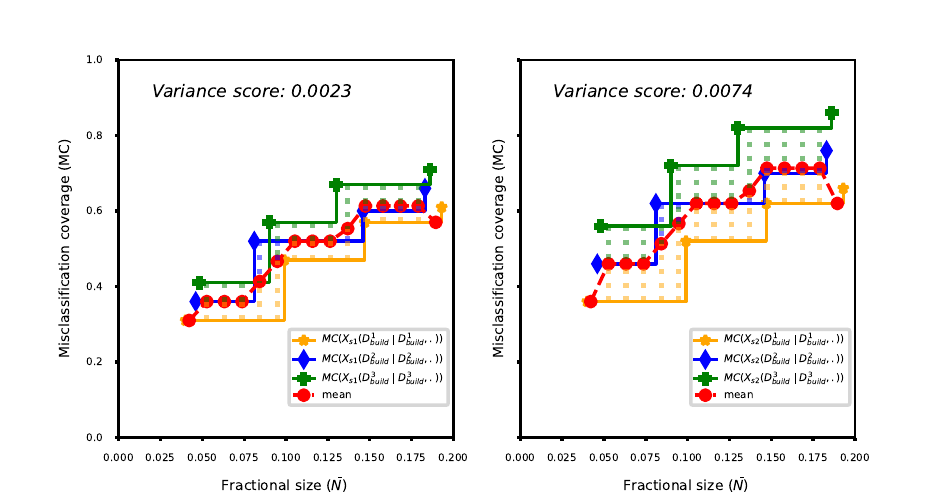}
    \caption{Stability values for two strategies $s_1$ (left) and $s_2$ (right): $V_{s_1}(\{\build{D^j}\}_{j=1}^3) $ and $ V_{s_2}(\{\build{D^j}\}_{j=1}^3)$. $s_1$ is more stable, as the MC gaps on the vertical axis are smaller.}
    \label{fig:stability}
\end{figure}

The strategy's stability is  measured in terms of variability in the misclassification coverage (MC) of the build set across the splits, over different budget levels. Figure \ref{fig:stability} depicts the stability of two different strategies. We can see that the left strategy is more stable because the curves (each representing one of the $Q$ splits) have lower variability on the vertical (MC) axis, across the horizontal axis (reflecting increasing budgets), compared to the right plot.

More formally, as in Section~\ref{subsec:AUC}, the MC step function on $\build{D}^j$ is defined by the ordered pairs $\{(n_i(\build{D^j}), m_i(\build{D^j}))\}_{i=1}^K$, representing the metrics $\bar{N}$ and $MC$ across budgets $\{b_i\}$.
Note that for a given budget $b_i$, the actual attention set fractional sizes $\{n_i(\build{D^j})\}_{j=1}^Q$, which are upper-bounded by $b_i$, may differ among the $M$ build samples, hence these curves may be slightly staggered in their orientation along the horizontal ($n=\bar{N}$) axis.  Our variability metric aggregates the vertical (error coverage $m=MC$) axis variance by interpolation along the horizontal axis: 

\begin{itemize}[leftmargin=*]
    \item Let $\hat{m}_i(\build{D^j})$ be the error coverage rate $MC(X_s(\build{D}^j\mid \build{D}^j,  b_i))$ interpolated along the step function if the attention set fractional size $\bar{N}$ equaled the budget $b_i$ (and not the actually observed $n_i$).  
    \item Define $Q_i,\:i=1,\dots, K$ as $Q_i=\splitatcommas{\{j\colon\: n_1(\build{D^j}) \leq b_i \leq n_k(\build{D^j}),\:j=1,\dots,Q\}}$; that is, $Q_i$ is the indices $\{j\}$ of the $Q$ splits for which budget $b_i$ falls in the observed range of attention set sizes (i.e., the indices of splits for which the interpolation to $\hat{m}_i$ is defined).  $Q_i=\emptyset$ typically for the highest budget $b_K$ (no split $j$'s $n_K(\build{D}^j)$ exactly equals it, even if it was satisfied) or for small budgets $b_i$ (no split  generated an attention set satisfying it).
    \item Let  $v_i=\textrm{variance}(\{\hat{m}_i(\build{D^j})\}_{j\in Q_i})$ be the variance of these values (the vertical spread when the horizontal axis value is fixed at $b_i$), considering only splits $Q_i$ for which they are defined. 
\end{itemize}

\noindent
Define the variability as the weighted average of the variances:
\begin{equation}
V_s(\{\build{D^j}\}_{j=1}^Q )=\frac{1}{\sum_{i=1}^K I(|Q_i| > 0)}\sum_{i=1}^K\frac{1}{\sqrt{|Q_i|}}v_i
\end{equation}

Lower variability $V_s$ indicates the strategy $s$ is more consistent across random splits, and hence that when we have results only on one dataset split, that it is more reliable.

\subsection{Aggregation of strategy metrics}
\label{subsec:topsis-score}

Multiple Criteria Decision Making (MDCM) involves selecting an optimal item from a set of (say, $m$) `alternatives' based on 
a combination of (say, $n$) numeric `criteria' 
measured on each.  An $m\times n$ matrix $\mathbf{Z}=\begin{bmatrix}z_{ij}\end{bmatrix}$, where $z_{ij}$ is the $\nth{j}$ criterion score on the $\nth{i}$ alternative, is formed.  A score $R_i\in[0,1]$ (which may weight criteria unequally) is calculated for each alternative $i$, 
with the optimal alternative having the highest score.  One such ranking method is TOPSIS (Technique for Order Preference by Similarity to Ideal Solution, \cite{TP20}; see also \cite{T2021}).

We use TOPSIS to determine the optimal strategy (the `alternatives') on a given dataset $D$, by combining its performance, generalizability, and stability metric values (the set of `criteria').  We give performance a weight of 0.6 and the others 0.2 each. Because TOPSIS scores are normalized to $[0,1]$, a strategy can be compared between different datasets--- on which the MCRs, and hence the difficulty of the attention set localization may differ---in a way that the metrics cannot.

\section{Experiment
\label{sec:experiment}}
To evaluate the different strategies presented in Section \ref{ssec:strategies}, we perform experiments on four open data sets (Section~\ref{subsec:exp_data}) and measure the metrics defined in Section \ref{sec:strategy_metrics}. Our experimental results, summarized in Section \ref{sec:results}, indicate that the FreaAI-slices based strategies are significantly better than the common baselines.

Given a source dataset, the experiment consists of (1) creating multiple random splits, as Section \ref{subsec:procedure} explains, and then (2) measuring the strategy metrics and averaging them for each strategy's attention sets across splits, as Section \ref{subsec:exp_measures} details.  

\subsection{Experimental procedure}\label{subsec:procedure}

Our experiments use the random forest (RF) classifier as the ML model $\mathcal{M}$.  In principle, any ML algorithm can be used with the strategies, though it must have a confidence-level measure to use the confidence-based strategy.

Given a dataset, we create $Q=10$ random splits, where each time the dataset is separated into mutually-exclusive training (70\%), build (15\%), and evaluation (15\%) sets, denoted $\splitatcommas{(\train{D}^1,\build{D}^1, \eval{D}^1), \dots, (\train{D}^Q,\build{D}^Q, \eval{D}^Q)}$.  The build and evaluation splits together comprise the test set.  
For each split $j=1,\dots,Q$, $\mathcal{M}$ is trained on $\train{D}^j$, and we obtain the label predictions $\hat{Y}$ on $\build{D}^j$ and $\eval{D}^j$. For the purposes of the experiment, the ground truth values $Y$ are known for  $\eval{D}^j$, though this may not be true in the field.
On each $\build{D}^j$, the strategies' attention rules are determined for each of the budgets $b_1,\dots,b_K$.  For FreaAI-slice-based strategies, the slices are determined on $\build{D}^j$ only.  The build splits are used to assess the sampling stability (Section~\ref{subsec:stability}), and each strategy-budget pair's attention rule is mapped to the corresponding evaluation dataset, to assess its generalizability (Section~\ref{subsec:general}).

\subsection{Experimental data}
\label{subsec:exp_data}
We experiment with four open datasets, available through the UCI Machine Learning Repository (\cite{UCI}). Table~\ref{tab:datasets} summarizes the datasets characteristics; the misclassification rate (MCR) is the average error rate of the classifier $\mathcal{M}$ on the test samples across random splits, for each dataset.
The datasets are: 
\begin{itemize}[leftmargin=*]
    \item \textbf{Adult}\footnote{Available at \url{https://archive.ics.uci.edu/ml/datasets/adult}}: A dataset extracted from US Census records, containing observations on individuals older than 16 years who worked at least one hour a week.  The target is a binary indicator of whether the person's income exceeded \$50,000.
    \item \textbf{Avila}\footnote{Original paper, \cite{DESTEFANO201899}; available at \url{https://archive.ics.uci.edu/ml/datasets/Avila\#}}: A dataset of layout features mainly related to the organization of the page and to the exploitation of the available space. It has been extracted from 800 images of the 'Avila Bible', an XII century giant Latin copy of the Bible. The target label represents a particular copyist. (A, B,  etc.).
    \item \textbf{Credit card defaults}\footnote{Original paper, \cite{YL2009}; available at \url{https://archive.ics.uci.edu/ml/datasets/default+of+credit+card+clients}}: A dataset of observations from October 2005 (during a financial crisis) of customers holding credit cards of a Taiwanese bank.  The target is a binary indicator of whether the person defaulted on a payment.
    \item \textbf{Electrical grid}\footnote{Original paper, \cite{AKJ18}; available at \url{https://archive.ics.uci.edu/ml/datasets/Electrical+Grid+Stability+Simulated+Data+}}: A dataset of simulated numeric input parameters to an electrical grid.  The target is a binary indicator of whether the system is stable or not.
\end{itemize}

\begin{table}
\renewcommand{\arraystretch}{1.0}
\begin{tabular}{| p{1.8cm}|p{1.4cm}|p{1.3cm}|p{0.9cm}|p{0.8cm}|p{0.8cm}|}
\hline

   & Categorical features & Numerical features & Records ($N$) & Unique labels & MCR ($\bar{M}$) \\ 
   \hline
Adult & 5 & 8 & 48,842 &2 & 0.15 \\
Cc default & 3 & 20 & 30,000 & 2 & 0.19 \\
Avila & 0 & 10 & 21,867 & 12 & 0.03 \\
Electrical grid & 0 & 12 & 10,000 & 2 & 0.12 \\
\hline
\end{tabular}
\caption{\label{tab:datasets} Properties of datasets used in the experiments.}
\end{table}

\setlength{\arrayrulewidth}{0.3mm}



\subsection{Experimental measurements}
\label{subsec:exp_measures}
We measure the strategy quality metrics by averaging their results on each of $(\build{D}^1, \eval{D}^1), \dots, (\build{D}^Q, \eval{D}^Q)$.  


\begin{enumerate}[leftmargin=*]
    \item \textbf{AUC averaged across the build splits}:
    \begin{equation}        \overline{AUC}_s=\frac{1}{Q}\sum_{j=1}^Q AUC_s(X(\build{D}^j \mid \build{D}^j))
    \end{equation}
    \item \textbf{Generalizability averaged across the splits}:
    \begin{equation}
        \bar{G}_s=\frac{1}{Q}\sum_{j=1}^Q    G_s(\build{D}^j, \eval{D}^j)
    \end{equation}
    \item \textbf{Stability across splits as in Section ~\ref{subsec:stability}}
  
\end{enumerate}

  Then, the TOPSIS score is calculated as explained in Section ~\ref{subsec:topsis-score}, inducing a ranking of the strategies for a given datasets.


\section{Experimental Results}
\label{sec:results}

\begin{figure}[h]
    \centering
    \includegraphics[width=\columnwidth]    {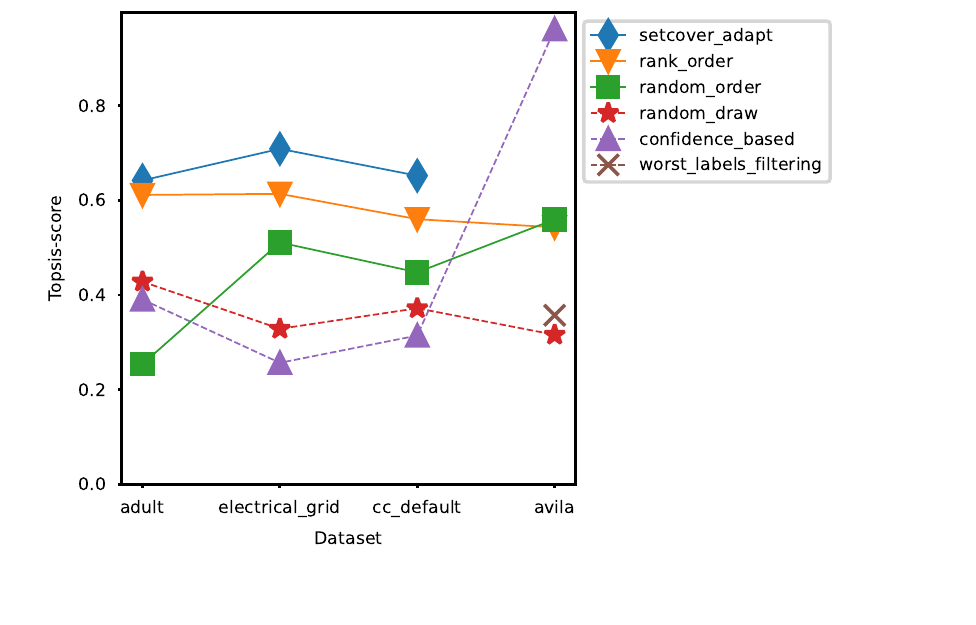}
    \caption{Values of the TOPSIS aggregate scores (Section~\ref{subsec:topsis-score}), averaged across $Q=10$ splits.}
    \label{fig:topsis-score-results}
\end{figure}

Figure~\ref{fig:topsis-score-results} shows the average build TOPSIS scores across splits, for each strategy and dataset.  Over the first three datasets, the FreaAI slice-based methods, set cover, rank-order, and random-order, have the best scores after aggregating the three quality metrics.  We note that the Avila dataset appears to be anomalous in that the confidence-based and random-order Frea strategies seem to have better performance than the others.  This may be due to Avila's very low MCR, which may be explained by the fact it was created through an extensive process of feature engineering that aims to optimize the prediction of the class. Therefore, ML models are able to more easily identify the relevant patterns and make more accurate predictions, in addition to having a higher level of confidence in their predictions. There are also too few errors for FreaAI's localization to generalize well.

\begin{figure}[h]

    \includegraphics[scale=0.9]
{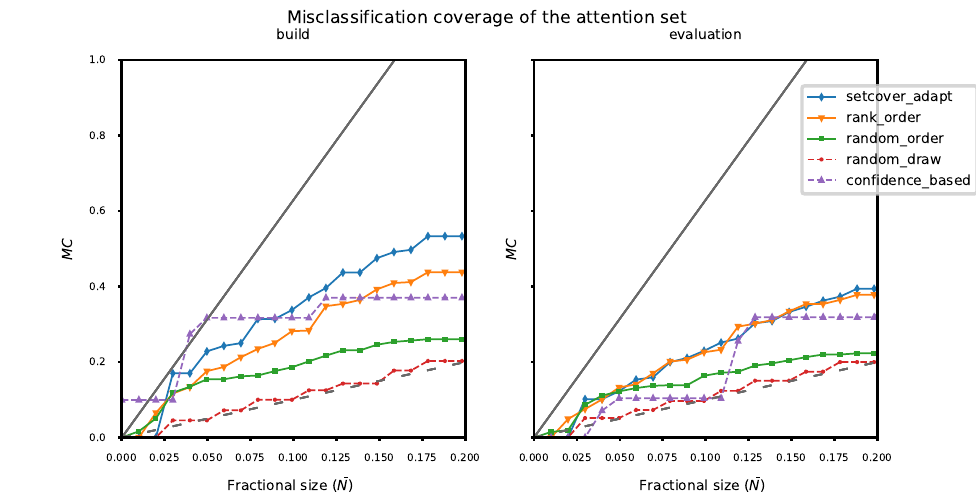}

    \caption{Values of the MC statistic (Section ~\ref{sec:attention_set}) for each strategy, averaged across the $Q=10$ build and evaluation splits from the Adult dataset.}
    \label{fig:experiment_mc__statistics}
\end{figure}

\begin{figure*}
    \centering
    \includegraphics[width=0.49\textwidth]
    {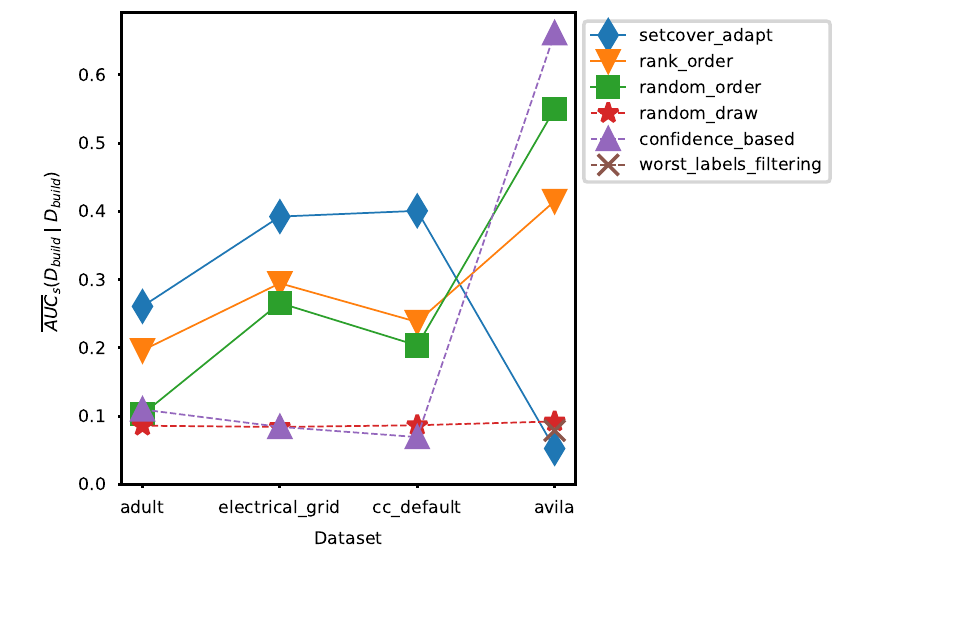}
    \includegraphics[width=0.49\textwidth]
    {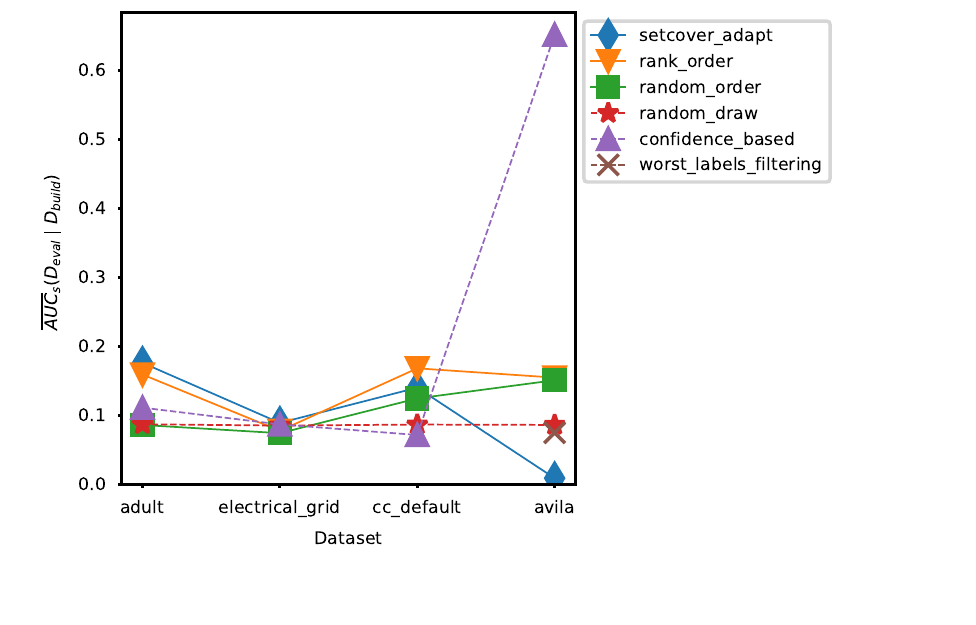}
    \includegraphics[width=0.49\textwidth]
    {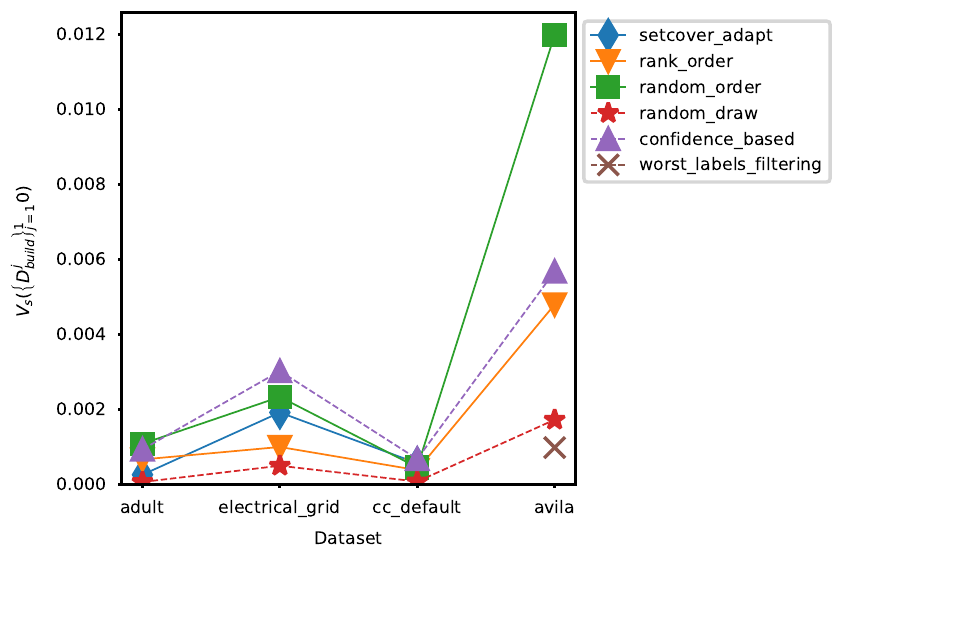}
    \includegraphics[width=0.49\textwidth]
    {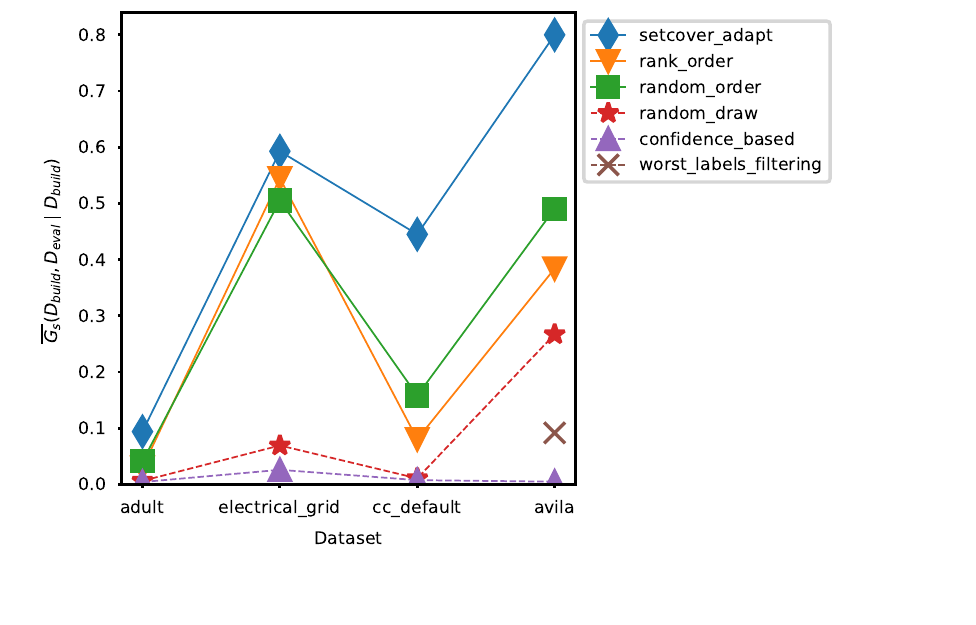}
    \caption{Values of quality metrics (Section~\ref{sec:strategy_metrics}) for each strategy, for each dataset, averaged across $Q=10$ splits.  \\Top row: AUC on build (left) and evaluation (right) datasets.  Higher values are better.  \\Bottom row: stability $V_s$ (left) on build datasets; generalizability (right) from build to evaluation datasets.  Lower values are better.}
    \label{fig:experiment_results}
\end{figure*}

Figure~\ref{fig:experiment_mc__statistics} shows the averaged $MC$ vs $\bar{N}$ plots of the attention sets found on the Adult dataset, across a range of budgets $b$ up to $B=0.20$. Similarly to the toy examples in Figure~\ref{fig:Attentionsets}, the error coverage should increase at a decreasing rate as the budget increases.  On this dataset, the slice-based strategy (set cover and rank-order) curves are higher than the others on the build dataset, but on the evaluation, the gap is lower than on the build, indicating that some overfitting may be happening.  These correspond to the top row of Figure~\ref{fig:experiment_results}, where these strategies have the highest AUC performance (except on the Avila dataset, which we noted is anomalous) on the build datasets, but are more similar to the other strategies on the evaluation data.  The higher build AUC indicates that the FreaAI-based strategies are generally more effective in identifying and addressing areas of model error than the baseline strategies.  Moreover, as noted in Section~\ref{subsub:strategies_overview}, Avila is the only dataset on which the worst-labels strategy is able to event create an attention rule, because the datasets only have two unique label values.

Another important aspect of a strategy is its ability to consistently identify model errors in its attention sets.  As outlined in Section ~\ref{subsec:stability}, stability is measured by analyzing the variations in misclassification coverage across different attention sets, obtained with a fixed budget and through multiple samples of the same dataset distribution.
When considering $Q$ random samples (here, splits) of the same dataset, the attention sets generated should demonstrate a stable and consistent level of error coverage (MC).
As shown in the bottom left of Figure ~\ref{fig:experiment_results}, the FreaAI-based strategies have better (lower $V_s$ values) or comparable stability to the baselines, particularly the confidence-based one.  The random observation draw strategy has good stability (due to its total randomness) but this aspect isn't useful in this case because the strategy itself is non-informative in localizing errors.

The last aspect of a strategy that we evaluate is how well it generalizes to unseen data. The bottom right of Figure ~\ref{fig:experiment_results} shows the average measure of generalizability for each strategy across all four datasets. The strategies based on FreaAI slices show poor generalizability, as they performed significantly better on the build dataset. This implies that the attention sets identified might suffer from over-fitting and generate attention rules that are overly specific. 
We note that the random order strategy simply combines the FreaAI slices in a random manner, and therefore provides and indication of the FreaAI slices performance, stability and generalizability. 

\section{Related work}\label{sec:related}

There are several software tools, such as Google's SliceFinder (\cite{SliceFinder}) and Microsoft's Error Analysis (\cite{ErrorAnalysis}), that operate similarly to FreaAI, using decision trees to isolate feature-value-based slices that contain error concentrations.  
However, because multiple slices would likely be needed to properly diagnose a model, it would be difficult for a user to effectively consolidate these slices into a single unit.  The attention rule method we propose does this automatically, at a user-specified budget $b$, and with appropriate analysis metrics (Section~\ref{sec:strategy_metrics}).  The attention rule approach specifically aims to avoid overloading the user with individual slice subsets.

A tool that performs a similar function to attention rules in the regression context is `Evidently'.  In \cite{Evidently} they illustrate a scatterplot of predicted (continuous) $\hat{y}_i$ vs true $y_i$, with a regression line; the 10\% of observations furthest from the line (over- or under-predictions) could constitute an attention set at $b=0.1$.  
However, because these large-error observations are not a priori localized by feature values (as FreaAI slices are), they may not necessarily be attributable to particular feature values, in which case the model may not need diagnosis, because there will always exist a group of 10\% of observations with largest errors. 

An idea similar to our attention rules is proposed by \cite{JRS1996}, albeit for diagnosis of hardware processors and not ML models on static datasets.  They consider branch prediction, where the processor must predict (ideally with low error) the future state of a machine while it processes a given current instruction.  In our terminology, each branch could be considered as a `slice' which maps to a set of instructions that reach it; a given set of branches could be seen as an attention rule.  They show that in a static case, the branches can be arranged from highest to lowest error; for each number $k$, considering the set of $k$ worst branches (an attention rule), $k$ can be plotted vs the cumulative share of mispredictions it contains, similar to Figure~\ref{fig:Attentionsets}.  A `low confidence set' of branches can be found statically by selecting either a target maximal number of branches or proportion of mispredictions (a sort of budget), but they propose dynamic methods as well.  While their approach is similar, the setting is different and the selected low-confidence branches may not have shared causes. We are interested in identifying problematic features that may suggest underlying causes.

We note also that in the case of slices, our approach has connections to the subgroup discovery (SD, also known as `association rules' modeling; see, e.g., \cite{HCGJ2010}).
An example of a subgroup rule is $(\splitatampersand{\{\textrm{INCOME}\geq \$100,000\}\:\&\:\{\textrm{OWN\_HOUSE}\in\{\textrm{True}\}\}})\rightarrow\{\textrm{DEFAULT}\in\{\textrm{False}\}\}$, where the antecedent (the part of the rule left of the arrow) specifies a subset observations (i.e., a slice rule), and the right side specifies a condition on feature of interest, for which the condition is likely to be true given the antecedent.  For instance, this rule, if found, implies that people with income above \$100,000 who own their home are unlikely to default on a loan.  In FreaAI slices, the target feature is specifically the binary indicator $Z=I(Y=\hat{Y})$,
and we are only interested in results where `$Z=\textrm{False}$' is more likely than average.  The quality of returned subgroups is often assessed by support, precision, and coverage metrics similar in nature to those in Section~\ref{sec:attention_set} for attention sets.  Often, competing SD algorithms are assessed on the individual subsets (e.g., comparing the highest-quality subset from each algorithm).  However, methods such as \cite{LK2012} propose search methods that balance rule diversity and exploration and result in fewer redundant subsets caused by significant overlap in observation membership between subsets.  Such techniques have a similar goal to, say, our set cover attention rule strategy (Section~\ref{subsub:FreaAIStrategy}), which likewise tries to build a union of slices that minimizes overlaps in observation coverage.


\section{Conclusion and Discussion}
\label{sec:conclusion}

In this work, we have presented the concepts of attention rules and sets, which are limited-size subsets of observations on which an ML classifier model's predictions are most likely to be mistakes.  These sets serve to isolate error-prone observations, which may be used to diagnose the model.  We presented several strategies, or algorithms, for identifying optimal attention sets, several based on FreaAI (\cite{ARZ2020}) feature slice-finding technology.  We also evaluated these strategies, in terms of their performance, sampling stability, and generalizability to unseen data, on multiple datasets.



\section{Data Availability
\label{sec:data_availability}}

We plan to release a compiled version of the algorithm code, as well as code for the experiments, in a publicly-available github repository.  The URL will be revealed later so as not to violate the double-blind review.

\bibliography{main}

\end{document}